\documentclass{applemlr}

\usepackage{amsmath}
\usepackage{enumerate}
\usepackage{algorithm}
\usepackage{algpseudocode}
\usepackage{amsfonts}
\usepackage{amsthm}
\usepackage{cleveref}
\usepackage{diagbox}
\usepackage{colortbl}
\usepackage{amssymb}
\usepackage{xspace}
\usepackage{wrapfig}
\usepackage{adjustbox}
\usepackage{tabularx}
\usepackage{booktabs}
\usepackage{mathtools}
\usepackage{tikz}
\usepackage{enumitem}
\usepackage{silence}
\usepackage{dsfont}
\usepackage[table]{xcolor}
\usepackage[dvipsnames]{xcolor}
\usepackage{multirow}
\usepackage{makecell}
\usepackage{xfakebold}
\usepackage{amsmath,amsfonts,bm}

\def\eqref#1{equation~\ref{#1}}
\def\1{\bm{1}}

\DeclareMathAlphabet{\mathsfit}{\encodingdefault}{\sfdefault}{m}{sl}
\SetMathAlphabet{\mathsfit}{bold}{\encodingdefault}{\sfdefault}{bx}{n}

\definecolor{textgray}{HTML}{6E6E73}
\usetikzlibrary{positioning, calc}
\usetikzlibrary{decorations.pathmorphing}

\makeatletter
\patchcmd{\wrong@fontshape}{\@gobbletwo}{}{}{}
\makeatother
\numberwithin{equation}{section}
\makeatletter
\AtBeginDocument{
  \urlstyle{sf}
  
}
\makeatother

\definecolor{light}{RGB}{125, 125, 125}
\crefname{tcb@cnt@pbox}{code}{code}
\Crefname{tcb@cnt@pbox}{Code}{Code}
\crefname{assumption}{assumption}{assumption}
\Crefname{assumption}{Assumption}{Assumptions}

\newtcolorbox[auto counter]{pbox}[2][]{
  colback=white,
  title=Code~\thetcbcounter: #2,
  #1,fonttitle=\sffamily,
  fontupper=\sffamily,
  arc=2pt,
  colframe=bgcolor,
  coltitle=fgcolor,
  colbacktitle=bgcolor,
  toptitle=0.25cm,
  bottomtitle=0.125cm
}

\makeatletter
\newcommand\applefootnote[1]{%
  \begingroup
  \renewcommand\thefootnote{}%
  \renewcommand\@makefntext[1]{\noindent##1}%
  \footnote{#1}%
  \addtocounter{footnote}{-1}%
  \endgroup
}
\makeatother

\definecolor{cverbbg}{gray}{0.90}

\newcommand{\y}{ \checkmark }
\newcommand{\n}{ }
\newcommand{\ar}{$\rightarrow$}

\title{A Small-Scale System for Autoregressive Program Synthesis Enabling Controlled Experimentation}

\author[*]{Russ Webb}
\author{Jason Ramapuram}

\affiliation{Apple}

\abstract{
\emph{What research can be pursued with small models trained to complete true programs?} Typically, researchers study program synthesis via large language models (LLMs) which introduce issues such as knowing what is in or out of distribution, understanding fine-tuning effects, understanding the effects of tokenization, and higher demand on compute and storage to carry out experiments. We present a system called Cadmus which includes an integer virtual machine (VM), a dataset composed of true programs of diverse tasks, and an autoregressive transformer model that is trained for under \$200 of compute cost. The system can be used to study program completion, out-of-distribution representations, inductive reasoning, and instruction following in a setting where researchers have effective and affordable fine-grained control of the training distribution and the ability to inspect and instrument models. Smaller models working on complex reasoning tasks enable instrumentation and investigations that may be prohibitively expensive on larger models. To demonstrate that these tasks are complex enough to be of interest, we show that these Cadmus models outperform GPT-5 (by achieving 100\% accuracy while GPT-5 has 95\% accuracy) even on a simple task of completing correct, integer arithmetic programs in our domain-specific language (DSL) while providing transparency into the dataset’s relationship to the problem. We also show that GPT-5 brings unknown priors into its reasoning process when solving the same tasks, demonstrating a confounding factor that prevents the use of large-scale LLMs for some investigations where the training set relationship to the task needs to be fully understood.
}

\metadata[Correspondence]{\sffamily Russ Webb: \url{rwebb@apple.com}; Jason Ramapuram (Work done while at Apple.): \url{jason@ramapuram.net}}
\date{\sffamily\today}

\begin{document}

\maketitle

\section{Introduction and Related Work}

The system presented here is called Cadmus\footnote{The name is inspired by the inductive logic card game Eleusis \cite{cardgame}, invented by Robert Abbott.  The mythohistorical connections between Eleusis and Cadmus are complex, though some claim Cadmus supplanted the family of Eleusis, others claim there was a familial relation.  Cadmus is typically credited with the founding of Thebes and the introduction of the Phoenician alphabet to the Greeks.\label{cadmusname}}.  Although Cadmus is capable of sequence labeling, subroutine use, vector operations, processing decorated integers, and algorithm induction tasks, the results presented here focus on a subset of Cadmus VM instructions:  literal values, basic integer math operations, integer comparisons, and program termination.

Many investigations, models, and simple math benchmarks \cite{hendrycks21mdata} in the literature overlap this work, from work on inductive program synthesis \cite{programsketches, bootstrappingsynthesis} to reasoning from examples to solve more examples in \cite{ARC}.  Some of the key goals in this work are 1) avoiding natural language training, 2) utilizing verifiable true programs (those that evaluate to true), 3) constructing a VM such that ANY sequence of instructions produces a fixed number of output values and no errors halt the execution of programs, 4) avoiding infinite loops by limiting recursion depth, 5) implementing sequence labeling and inductive program synthesis as a by-product of training on true programs, 6) ensuring program structure is suitable for multi-task causal language modeling, and 7) avoiding tokenization complexities by having a one-to-one, fixed tokenization for all VM instructions.  Notably, Nova \cite{nova} and ALMOND \cite{almond} present models trained on auto-regressive low-level instructions.  Similar to \cite{tiips}, we train and test on sampled programs from a domain specific language (DSL), however unlike the TIIPS which trains on a string manipulation DSL, Cadmus is trained via sampling from a post-fix, concatenative, integer computation and sequence labeling domain. The major differences in approaches are highlighted in Table \ref{tab:compare}.

A key question for any model training is what data to train on.  Most auto-regressive models for program synthesis are trained on a selection of syntactically correct programs, or those that are deemed to be typical.  In this work, the model is trained on true-programs, which are those that produce one or more true values.  All Cadmus VM instructions are non-variadic and there are two false values: zero and not-a-number (NAN).  Instructions that receive NAN as an input produce NAN for all outputs.  In this way, any sequence of instructions is an executable program that is either a true-program or a false-program (one that produces no values or some false values).  Under this approach, the model is trained via next token prediction to approximate the distribution of true-programs.  Cadmus programs are shown as a sequence of single character mnemonic instructions in \texttt{square brackets} and are terminated by `.'. Given the postfix VM instructions in Table \ref{tab:instructions}, the program \texttt{[34+7=.]} (i.e. $3+4=7$) is true while \texttt{[34+8=.]} (i.e. $3+4=8$) is false.  Note that false-programs can be negated into true-programs if they produce false rather than NAN, for example \texttt{[34+8=!.]} (i.e. $3+4 \neq 8$) is a true-program.

The model presented here is trained on randomly sampled true-programs from multiple templates, which are designed to enable the model to progressively learn the characteristics of true-programs.  The dataset composition with respect to these templated sub-sets of programs are listed in Table \ref{tab:subproblems} with pseudo-code since only a subset of the Cadmus instructions are detailed here due to space constraints.  The current Cadmus model, Cadmus-280M-80M-v1 (indicating the number of parameters, 280M, and the number of samples in the training set, 80M), is an 18 layer decoder-only transformer with the following configuration: 65 vocabulary size (the number of VM instructions), 1280 embedding dimension, 20 heads, 3600 MLP hidden size, GELU activation.  Training used  Adam (lr=1e-4 cosine schedule and betas: [0.9, 0.95]) for 300k steps with batch size 1024 on eight H100s.

\begin{table}[!ht]
\centering
\scriptsize
\caption{Comparison of works on transformers and related models trained on assembly code or short equation/algorithmic tasks. The table lists the following: \textbf{(Trans.)} the type of transformer used: E=encoder and D=decoder , \textbf{(Exec)} if the model can execute programs, \textbf{(Vals)} types of values  operated on L=list, I=integer, F=float, S=string, T=tokens, ASM=assembly string, \textbf{(DSL)} if the model learns a fixed, low level DSL, \textbf{(NL)} if the model is also trained on natural language, \textbf{(ProgSyn)} if the model does program synthesis, \textbf{(Induction)} if the model can inductively determine programs from examples, \textbf{(Label Seq)} if sequence labeling is a focus of the model, \textbf{(T|F)} if the model is designed to label program as true or false.}
\label{tab:compare}

\begin{tabular}{|l|c|c|c|c|c|c|c|c|c|}

\hline
%\textbf{}                                     & Q2     & Q3   & Q4     & Q5   & Q6 & Q7      & Q8        & Q9        & Q10 \\
\textbf{Work}                                  & Trans. & Exec & Vals   & DSL  & NL & ProgSyn & Induction & Label Seq & T|F \\
\hline
Cadmus (this work)                             & \ar D     & \y   & IS     & \y   & \n & \y      & \y        & \y        & \y  \\
\hline
TIIPS (2025) \cite{tiips}                      & E\ar D     & \y   & LIS    & \y   & \n & \y      & \y        & \n        & \n  \\
\hline
ALMOND (2025) \cite{almond}                    & \ar D      & \n   & ASM    & \y   & \n & \n      & \n        & \n        & \n  \\
\hline
Nova (2025) \cite{nova}                        & \ar D      & \n   & ASM    & \y   & \n & \y      & \n        & \n        & \n  \\
\hline
Tracr (2023) \cite{lindner2023tracr}           & \ar D      & \y   & T      & RASP & \n & \n      & \n        & \y        & \y  \\
\hline
Exedec (2023) \cite{exedec}                    & E\ar D     & \y   & LIS  & \y   & \n & \y      & \y        & \n        & \n  \\
\hline
Looped Transformer (2023) \cite{looped23}      & \ar D      & \y   & IT    & \n   & \y & \n      & \n        & \y        & \y  \\
\hline
Math Reasoning (2019) \cite{mreason19}         & E\ar D, \ar D  & \n & IF  & \n   & \y & \n      & \n        & \n        & \y  \\
\hline
NALU (2018) \cite{trask2018neural}             & \n         & \y   & IF    & \n   & \n & \n      & \n        & \n        & \n  \\
\hline
Neural GPUs (2016) \cite{kaiser2016neural}     & \n         & \y   & I      & \n   & \n & \n      & \y        & \n        & \n  \\
\hline
Learning to Execute (2014) \cite{learnexec14}  & \n         & \y   & I      & \n   & \n & \n      & \n        & \y        & \y  \\
\hline
\end{tabular}
\end{table}

%\vspace{-8pt}

\renewcommand{\arraystretch}{1.2}
\begin{table}[!ht]
  \begin{center}
    \caption{The subset of VM instructions used for the number comparison tasks are shown with their stack effect and the alternate (Alt) form used to test for robust LLM responses.}
    \label{tab:instructions}

    \begin{tabular}{r|l|l|||r|l|l} 
      Instruction & Stack Effect & Alt. Form & Instruction & Stack Effect & Alt. Form \\
      \hline
    \texttt{0} & ( $\rightarrow$ 0) & \texttt{-}  & \texttt{+} & (a b $\rightarrow$ a+b) & \texttt{*}  \\
    \texttt{1} & ( $\rightarrow$ 1) & \texttt{[}  & \texttt{-} & (a b $\rightarrow$ a-b) & \texttt{/} \\
    \texttt{2} & ( $\rightarrow$ 2) & \texttt{\_} & \texttt{*} & (a b $\rightarrow$ a*b) & \texttt{\%}   \\
    \texttt{3} & ( $\rightarrow$ 3) & \texttt{+}  & \texttt{/} & (a b $\rightarrow$ a//b) & \texttt{)} \\
    \texttt{4} & ( $\rightarrow$ 4) & \texttt{!}  & \texttt{\%} & (a b $\rightarrow$ a\%b) & \texttt{\}} \\
    \texttt{5} & ( $\rightarrow$ 5) & \texttt{\#} & \texttt{x} & (a b $\rightarrow$ max(a, b)) & \texttt{L} \\
    \texttt{6} & ( $\rightarrow$ 6) & \texttt{9}  & \texttt{n} & (a b $\rightarrow$ min(a, b)) & \texttt{b} \\
    \texttt{7} & ( $\rightarrow$ 7) & \texttt{1}  & \texttt{<} & (a b $\rightarrow$ 1 if a<b else 0) & \texttt{?} \\
    \texttt{8} & ( $\rightarrow$ 8) & \texttt{7}  & \texttt{>} & (a b $\rightarrow$ 1 if a>b else 0) & \texttt{\$} \\
    \texttt{9} & ( $\rightarrow$ 9) & \texttt{\^} & \texttt{=} & (a b $\rightarrow$ 1 if a==b else 0) & \texttt{\~} \\
    \texttt{.} & ( $\rightarrow$ )  & \texttt{.}  & \texttt{!} & (a $\rightarrow$ 0 if a else 1) & \texttt{\&} \\
    \end{tabular}
    
  \end{center}

\end{table}

\section{Experimental Results}

These sections are intended to provide a sampling of the types of experiments that are possible using the Cadmus system to configure training, testing, and instrumentation in a knowable, verified setting.  Many more experiments remain to be done including those aimed at program understanding and inductive reasoning from examples.  The model used here is the 280M parameter model detailed in the introduction; the model is trained on the data mixture in Table \ref{tab:subproblems}.
\renewcommand{\arraystretch}{1.2}
\begin{table}[!ht]
  \begin{center}
        \caption{Sub-problems sampled from random templates make up the Cadmus training set.  Only the first four sub-problem types are presented in this work, though the model is trained with the all true-programs indicated in the table.  Programs shown in \texttt{square brackets} are the actual programs that Cadmus processes; in simple cases the pseudo-code is given as well.  The validation accuracy of the model on 2k samples from each sub-problem is also provided.  Accuracies below 1.0 on more complex, compositional tasks highlight remaining challenges in program generalization.}
    \label{tab:subproblems}

    \begin{tabular}{|l|l|c|l|l|} 
\hline
      \textbf{Name} & \textbf{\makecell{Val. \\Acc.}} & \textbf{Description} & \textbf{Samples} & \textbf{Examples}  \\
\hline
      \makecell[l]{basic \\math} & 1.0 & \makecell{simple \\math calculations} & 10M & \makecell[l]{$1+2 > 0$ \\ \texttt{[12+0>]}}  \\
\hline
      equality & 1.0 & \makecell{numerical \\ equality} & 10M & \makecell[l]{$9*(4-1) = 5*5 + 2$ \\ \texttt{[941-*55*2+=]}}  \\
\hline
      < > & 1.0 & \makecell{greater than \\and less than} & 10M & $9*(4-1) > 5*5 + 1$  \\
\hline
      \makecell[l]{sub- \\routines} & 1.0 & \makecell{calling \\subroutines} & 10M & $f(x)=x*(4\!-\!1); f(9) < 5*5 + 1$ \\
                   &     &                                  &     & $f(9) = 5*5 + 2; f(x)=x*(4\!-\!1);$ \\
\hline
      random & 0.92 & \makecell{random true-programs} & 200k & \texttt{[8Bt-Z?Ex+'] [u2;\&c\$;b]}  \\
\hline
      \makecell[l]{basic \\sequences} & 1.0 & \makecell{simple \\ sequence operations} & 10M 
      & x=(1,2,3); x[0]==1;  \\
\hline
      \makecell[l]{advanced \\sequences} & 1.0 & \makecell{complex \\ sequence operations} & 5M 
      & x=(1,2,3); len(select(x>1))==2;  \\
\hline
      \makecell[l]{apply \\labels} & 1.0 & \makecell{label a sequence} & 10M & f(x)=len(x)>1; x=(1,2,3); f(x);  \\
\hline
      \makecell[l]{complete \\algorithm} & 0.96 &  \makecell{give examples \\inductively synthesize \\a program} & 15M 
      & \makecell[l]{f((1,2,3))=1; f((2,3))=0; \\f((7,))=0; f((1,2,3,4))=1; \\f(x)=len(x)>2;} \\
\hline
    
    \end{tabular}
    
  \end{center}

\end{table}

\subsection{Comparing Different Program Producing the Same Values}

\begin{figure}[!ht]
  \centering
    \includegraphics[width=0.75\textwidth]{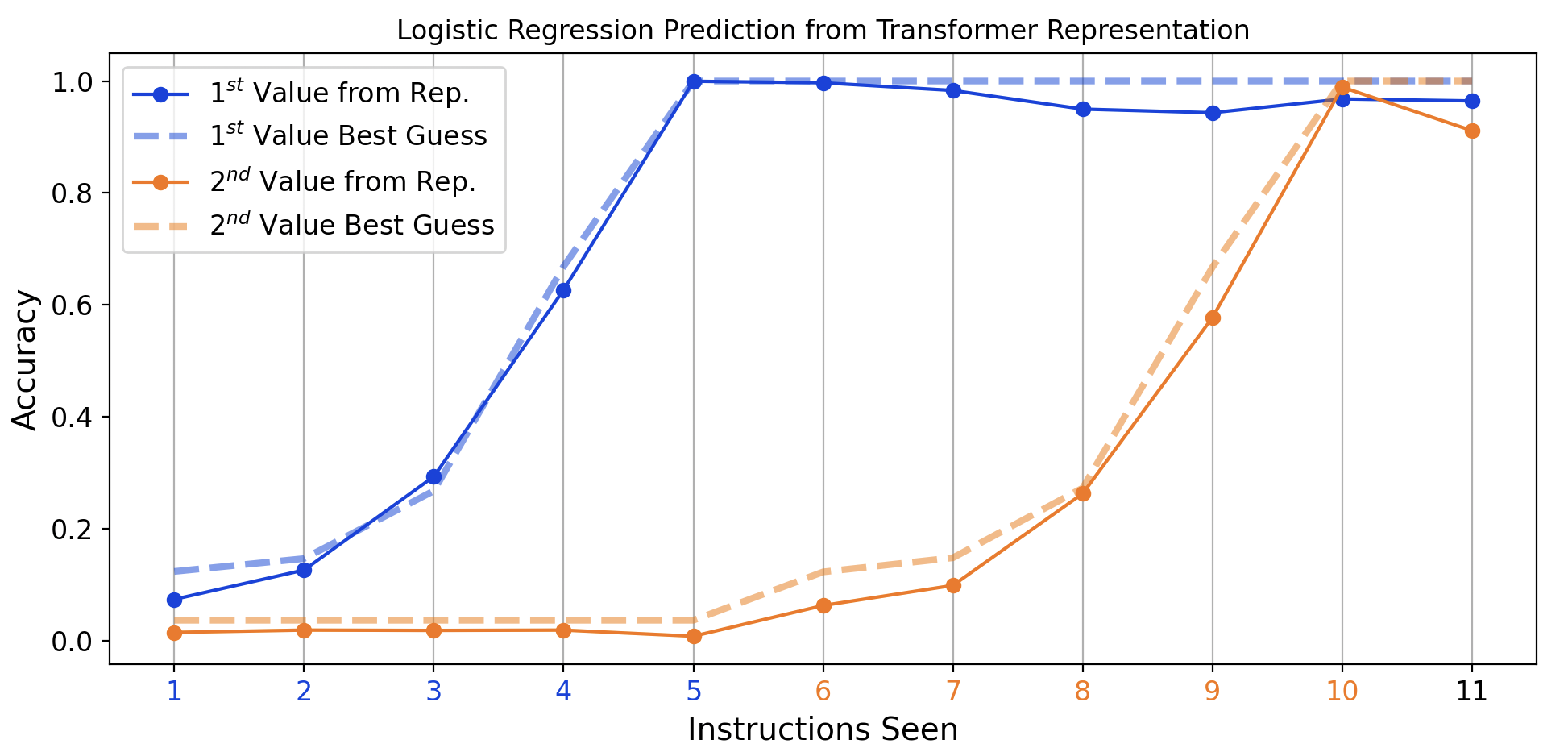}
\caption{The accuracies of logistic regression prediction of the two numbers computed during the number comparison task from the final transformer block representations at each token is shown.  Tokens 1--5 are computing the first number (blue), while the 6--10 tokens compute the second number; finally, token 11 is the comparison instruction.  The dashed lines show the accuracy obtained by guessing the majority answer in the dataset given the number of instructions seen.}
\label{fig:grid}
\end{figure}

Experimental results are presented to illustrate Cadmus and the value of small-scale, controlled, and verifiable experiments.  In Figure \ref{fig:grid}, programs which compute two numbers and compare them are tested using the Cadmus model and GPT-5 \cite{gpt5}.  For the Cadmus model, the \texttt{argmax} of the next token is used to predict the correct number comparison.  The dashed square is the in-distribution range in the Cadmus training set, for which the Cadmus model is shown to out-perform GPT-5 instructed to complete the same programs (the full LLM instructions are found in the appendix; also note that GPT-5 is run in batch mode where the output tokens gate reasoning, see Figure \ref{fig:acc_vs_tokens}).  Interestingly, redefining the symbols for the Cadmus instructions to be unrelated and in some cases intentionally confusing (for instance `9' pushing 6 onto the stack), effectively breaks all the GPT-5 models even though the updated instructions specify what each  symbol means.  This change shows that GPT-5 is utilizing some learned bias about what symbols mean to accurately model the VM.  These types of uncontrollable factors influence experimental outcomes and prevent a full understanding of an LLM's reasoning capabilities.  Further work on open models is needed to understand LLM performance on such queries.  The diagonal line averaging 46\% accuracy Cadmus 280M AR results are for out-of-training-distribution values and indicate that the generalized representations are not precise enough for equality comparisons when the model has never seen the values in training; again, this is an interesting area for further investigation.

\begin{figure}[htbp]
  \centering
    %\fbox{
    %                           left  bot   right top
    \includegraphics[clip, trim=0.5cm 3.8cm 0.5cm 3.8cm, width=0.85\textwidth]{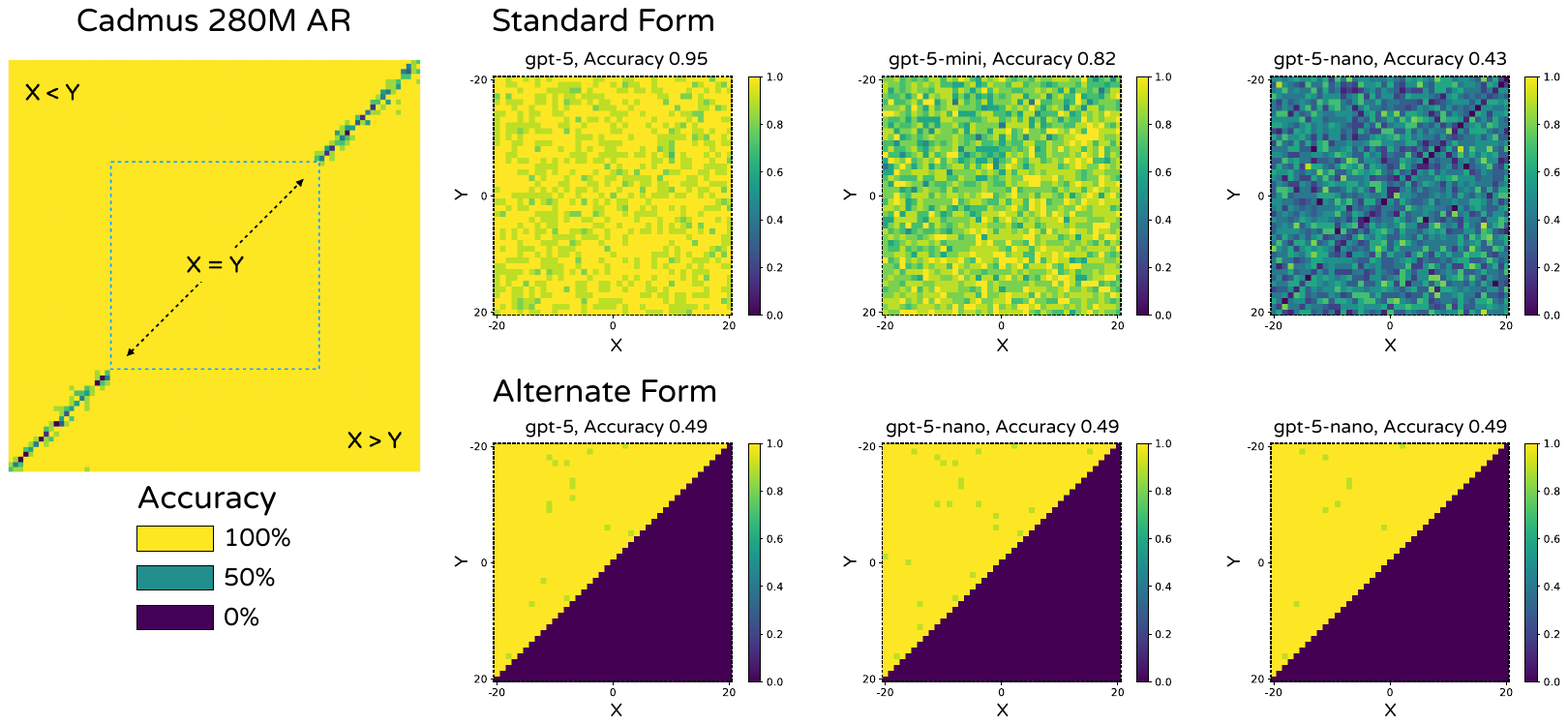}
    %}
\caption{The accuracy of predicting the correct comparison X and Y values is shown for in and out of distribution for the Cadmus-280M-80M-v1 model. The domain in-distribution for Cadmus-280M-80M-v1 is shown for GPT-5 models when given the instructions shown the Appendix A and 2k output tokens.  When provided with correct instruction describing the instructions in the alternate form, the GPT models are not able to provide useful responses using 2k output tokens (details in Appendix Figure \ref{fig:acc_vs_tokens}). In all cases, the accuracies reported are obtained by testing 10 validation programs per pixel (i.e. for each X,Y value pair).}
\label{fig:acc_per_instruction}
\end{figure}

\subsection{Building Representations of Computed Values}
A result of investigating how the Cadmus model builds numerical representations is shown in Figure \ref{fig:acc_per_instruction}.   Logistic regression on the final transformer layer's representation closely approximates the accuracy of guessing the most common result from the dataset.  The numerical comparison dataset is drawn from the set of all possible five instruction programs to compute a single value, and because the model is trained on a larger set of programs its accuracy is not strictly governed by the optimal guess.  Interestingly there is a decrease in the accuracy while the second number is calculated and a recovery of accuracy when both numbers have been computed at step 10.  Possibly the representation is less separable as the second value is calculated.  Finally, after the comparison operation at step 11 the two numbers are still represented (with a slight accuracy decrease) even though both numbers have been erased from the VM.  This retention of information deserves further investigation and may be due to either there being no need to instantly forget the numbers, or the past computation (and hence values) still having some predictive power within the training distribution of programs.

\subsection{Accuracy vs. Maximum Output Tokens}

\begin{figure}[htbp]
  \centering
    \includegraphics[width=0.8\textwidth]{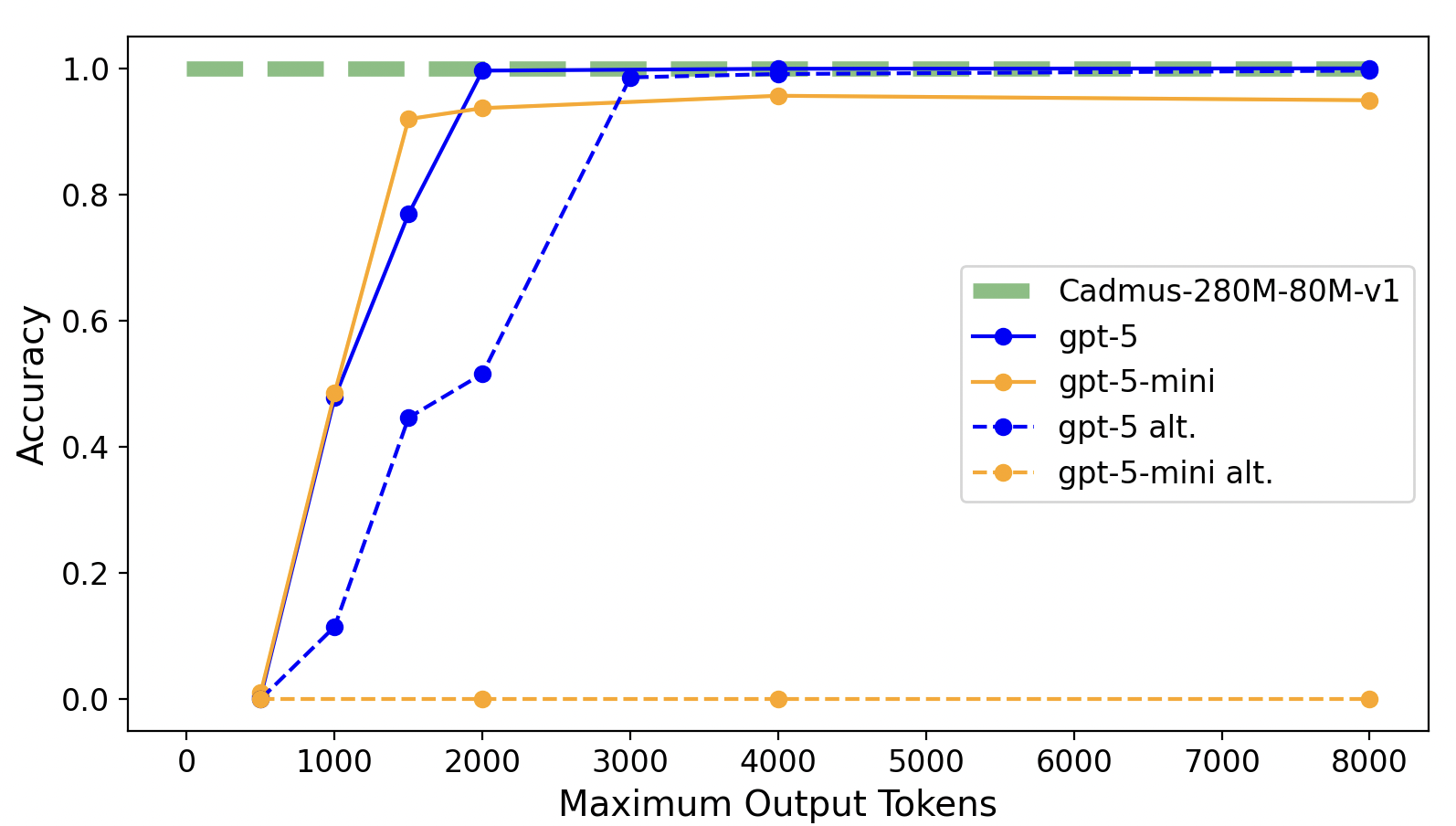}
\caption{The accuracy of predicting the correct comparison X and Y values is shown while varying the maximum output tokens (the Cadmus model produces an answer for each sample via predicting the next token).}
\label{fig:acc_vs_tokens}
\end{figure}

Here we look at how the accuracy changes versus maximum output tokens.  Figure \ref{fig:acc_vs_tokens} shows that GPT-5 uses fewer tokens in solving the standard Cadmus programs than the ones presented in the alternate form that is not aligned with typical expectations (see Alt. Form in Table \ref{tab:instructions}).  These additional tokens needed for the LLM to solve the task\footnote{The final accuracy is less than that of the standard Cadmus program form.} are evidence that the priors absorbed in training by LLM are a confounding factor in understanding how models approximate or implement reasoning. Several observations are worth noting: 1) the alternate form of programs is not solved by increasing token count, 2) a minimum token count is needed to solve the normal program encoding, and 3) gpt-5-mini does not improve with higher token counts indicating that is may be more reliant on learned instruction biases for effective reasoning.  For each data point, 1681 programs are tested (one for each X,Y value in the inclusive range [-20, 20].

\section{Conclusion}

The framework for Cadmus will be released and includes the VM verifier, templates for random program sampling, full 65 instruction set specification, examples of sequence labeling and program induction, current model, and full training and dataset code.  It is hoped that these facilities will encourage investigations, even with limited resources, into topics such as curriculums, discrete diffusion, numerical representations, inductive reasoning, and out-of-distribution generalization.

\section{Acknowledgements}
Thanks to many great, insightful collaborators for discussions and review, including Dan Busbridge, Eeshan Gunesh Dhekane, Amitis Shidani, Barry Theobald, Samy Bengio, Omid Saremi, Vimal Thilak, and Jerremy Holland, all of whom make research more engaging and productive.

% \section{Regular Font Tests}
% Regular text: Testing -- (en-dash) and --- (em-dash) ligatures here.
% Also testing \textbf{bold text: Testing -- (en-dash) and --- (em-dash) ligatures here.}

% \section*{Direct Font Commands}
% {\fontfamily{sfpro}\selectfont Regular SF Pro: Testing -- and --- ligatures. Question mark: ?}

% {\fontfamily{sfpro}\bfseries\selectfont Bold SF Pro: Testing -- and --- ligatures. Question mark: ?}

% {\fontfamily{sfpro}\itshape\selectfont Italic SF Pro: Testing -- and --- ligatures. Question mark: ?}

% \applefootnote{ \textcolor{textgray}{\sffamily Apple and the Apple logo are trademarks of Apple Inc., registered in the U.S. and other countries and regions.}}

\end{document}